\documentclass[conference,a4paper]{IEEEtran}
\IEEEoverridecommandlockouts
\usepackage{tabularx}
\usepackage{booktabs}
\usepackage{url}

\usepackage{amsmath,amssymb,amsfonts}
\usepackage{algorithmic}
\usepackage{textcomp}
\usepackage{xcolor}
\usepackage{listings}
\usepackage{adjustbox} 

\usepackage[utf8]{inputenc}
\usepackage[T5]{fontenc}
\usepackage[english]{babel}
\usepackage{csquotes}
\usepackage{tikzsymbols}

\usepackage{booktabs}
\usepackage{tabularx} 
\usepackage{stfloats}      
\usepackage{cuted}         
\usepackage{balance}       
\usepackage{placeins}      
\usepackage{needspace}      
\usepackage{caption}
\usepackage{subcaption} 
\usepackage{float}      
\usepackage{placeins}   

\usepackage{orcidlink} 

\usepackage{microtype} 
\usepackage{enumitem}
\setlist{nosep,leftmargin=*}

\usepackage{caption}
\usepackage{microtype} 
\microtypesetup{protrusion=false,expansion=true}

\usepackage{url}
\expandafter\def\expandafter\UrlBreaks\expandafter{%
  \UrlBreaks\do\.\do\-\do\_\do\/\do\?}

\makeatletter
\g@addto@macro\normalsize{%
  \setlength\abovedisplayskip{8pt plus 2pt minus 4pt}%
  \setlength\belowdisplayskip{8pt plus 2pt minus 4pt}%
  \setlength\abovedisplayshortskip{6pt plus 2pt minus 3pt}%
  \setlength\belowdisplayshortskip{6pt plus 2pt minus 3pt}%
}
\makeatother

\def\BibTeX{{\rm B\kern-.05em{\sc i\kern-.025em b}\kern-.08em
    T\kern-.1667em\lower.7ex\hbox{E}\kern-.125emX}}

\usepackage[noabbrev,capitalise]{cleveref}

\makeatletter
\def\@IEEEsectpunct{.\ \,} 
\def\@IEEEsectpunct@space{.\ } 

\def\section{\@startsection{section}{1}{\z@}%
  {0.6\baselineskip plus 0.2\baselineskip minus 0.2\baselineskip}
  {0.25\baselineskip}
  {\normalfont\bfseries\centering\uppercase}}

\def\subsection{\@startsection{subsection}{2}{\z@}%
  {0.5\baselineskip plus 0.2\baselineskip minus 0.2\baselineskip}
  {0.2\baselineskip}
  {\normalfont\bfseries\itshape}}
\makeatother

\usepackage{amsmath,amssymb,mathtools}

\makeatletter
\let\RIVForigthebibliography\thebibliography
\renewcommand{\thebibliography}[1]{%
  \RIVForigthebibliography{#1}%
  \fontsize{6.65pt}{7.0pt}\selectfont
  \setlength{\itemsep}{0pt}%
  \setlength{\parsep}{0pt}%
}
\makeatother

\begin{document}

\title{A Verifier-Guided Explainable Reasoning Framework with Gold-Anchored QLoRA, Task-Aware Mixture-of-Experts, and Group-Relative RLVR}

\author{
\IEEEauthorblockN{
Thi Kim Trang Vo$^{1,3}$,
Nam Tien Le$^{2,3}$,
Thi Kim Nguyet Vo$^{4,5}$,
Minh Khang Tran$^{1,3}$,
Duy Phuong Tran$^{1,3}$
}

\IEEEauthorblockA{
$^{1}$University of Information Technology (UIT), Ho Chi Minh City, Vietnam\\
$^{2}$Ho Chi Minh City University of Technology (HCMUT), Vietnam\\
$^{3}$Vietnam National University, Ho Chi Minh City, Vietnam\\
$^{4}$University of Economics Ho Chi Minh City (UEH), Vietnam\\
$^{5}$Viet Nam -- The Netherlands Programme (VNP), Vietnam
}
}

\maketitle
\thispagestyle{empty}
\pagestyle{empty}

\begin{abstract}
\normalfont
Large language models (LLMs) show strong reasoning ability, but their
explanations can remain inconsistent, weakly grounded, or difficult to
verify. We propose a \textbf{verifier-guided explainable reasoning framework}
for transparent educational question answering that combines
\textbf{gold-anchored QLoRA}, \textbf{task-aware symbolic routing}, and
\textbf{group-relative RLVR}. Qwen2.5-3B-Instruct is first adapted with
field-weighted QLoRA supervision anchored to authoritative answers. A
lightweight router then assigns logic problems to a \textbf{FOL/Z3 verifier}
and physics problems to a \textbf{formula- and unit-aware symbolic solver}.
Verifier feedback is further used to support candidate evaluation,
self-revision, and reward construction during RLVR. Candidate responses are
evaluated along three complementary dimensions: \textbf{P1} for answer
correctness, \textbf{P2} for evidence or unit consistency, and \textbf{P3}
for reasoning depth and explainability. At inference, gold-free
self-consistency aggregates multiple candidate responses before an optional
question-only physics verifier performs conservative system-level correction.
On 438 held-out examples, RLVR increases P3 from \textbf{50.68\%} to
\textbf{72.20\%}, while hybrid P1 remains approximately stable at
\textbf{55.94\%}. Self-consistency improves model-only P1 from
\textbf{48.86\%} to \textbf{50.23\%}, with symbolic verification providing
the remaining hybrid gain. These results indicate that RLVR primarily
strengthens explicit reasoning structure, while symbolic verification
complements the neural policy by improving answer reliability at the system
level.

\vspace{1em} 
\textbf{Our code implementation is available at:}\\ \url{https://github.com/VoThiKimTrang06101997/Explainable-xAI/tree/master/One-Shot-RLVR}\\


\end{abstract}

\section{Introduction}
\label{sec:introduction}

Large language models (LLMs) have achieved strong multi-step reasoning
performance through chain-of-thought (CoT) prompting and instruction tuning
\cite{wei2022cot,qwen25}. However, in scientific and educational question
answering, a plausible explanation is not necessarily trustworthy. A correct
answer may still rely on an invalid inference, an inappropriate physical
formula, unsupported evidence, or inconsistent units. Thus, reliable
reasoning requires both answer correctness and inspectable intermediate
reasoning.

Prior work improves reasoning through self-consistency
\cite{wang2023selfconsistency}, process supervision
\cite{lightman2023verify}, and scientific-logicality evaluation
\cite{yu2026scientificlogicality}. Reinforcement learning with verifiable
rewards (RLVR) further enables optimization from automatically checkable
signals. DeepSeekMath introduced Group Relative Policy Optimization (GRPO)
\cite{shao2024deepseekmath}, while One-Shot-RLVR demonstrated reasoning
improvements with limited supervision \cite{wang2025oneshotrlvr}.
Nevertheless, answer-only rewards cannot distinguish an accidentally correct
prediction from a well-supported solution.

This challenge is closely aligned with the \emph{2nd International XAI
Challenge for Transparent Educational Question-Answering (EXACT 2026)},
organized with IJCNN 2026 \cite{exact2026}. Its heterogeneous tasks require
different verification mechanisms: logic questions require premise-based
entailment checking, whereas physics questions additionally require formula,
numerical, and unit consistency.

To address these requirements, we propose a
\emph{Verifier-Guided Explainable Reasoning Framework with Gold-Anchored
QLoRA, Task-Aware Mixture-of-Experts, and Group-Relative RLVR}.
Qwen2.5-3B-Instruct \cite{qwen25} is adapted using QLoRA
\cite{dettmers2023qlora}, following LoRA \cite{hu2021lora}. Training targets
are anchored to authoritative answers, while field-weighted supervision
emphasizes answers, evidence, and units. Unlike conventional neural MoE
architectures \cite{fedus2022switch,jiang2024mixtral}, our lightweight router
dispatches tasks to external symbolic experts: FOL/Z3 verification for logic
\cite{demoura2008z3} and a formula- and unit-aware solver for physics.

We evaluate three complementary dimensions: \textbf{P1} for final-answer
correctness, \textbf{P2} for evidence or unit consistency, and \textbf{P3}
for reasoning depth and explainability. Verifier feedback supports candidate
evaluation, self-revision, and group-relative RLVR, while inference combines
gold-free self-consistency with optional conservative physics verification.
On 438 held-out examples, RLVR increases P3 from $50.68\%$ to $72.20\%$,
while hybrid P1 remains approximately stable at $55.94\%$, indicating that
RLVR primarily strengthens explicit reasoning structure while symbolic
verification contributes complementary system-level reliability.

Our contributions are mainly:
\begin{itemize}
    \item \textbf{Gold-anchored QLoRA:} a field-weighted adaptation strategy
    that preserves authoritative answers and structured evidence before RLVR.

    \item \textbf{Task-aware neuro-symbolic verification:} lightweight routing
    to FOL/Z3 logic verification and formula- and unit-aware physics
    verification.

    \item \textbf{Verifier-guided P1/P2/P3 RLVR:} a framework that separates
    answer correctness, evidence/unit consistency, and reasoning depth while
    distinguishing neural-policy performance from hybrid-system gains.
\end{itemize}


\section{Method}
\label{sec:method}

\subsection{Framework Overview}

\begin{figure*}[t]
    \centering
    \includegraphics[width=0.97\textwidth]{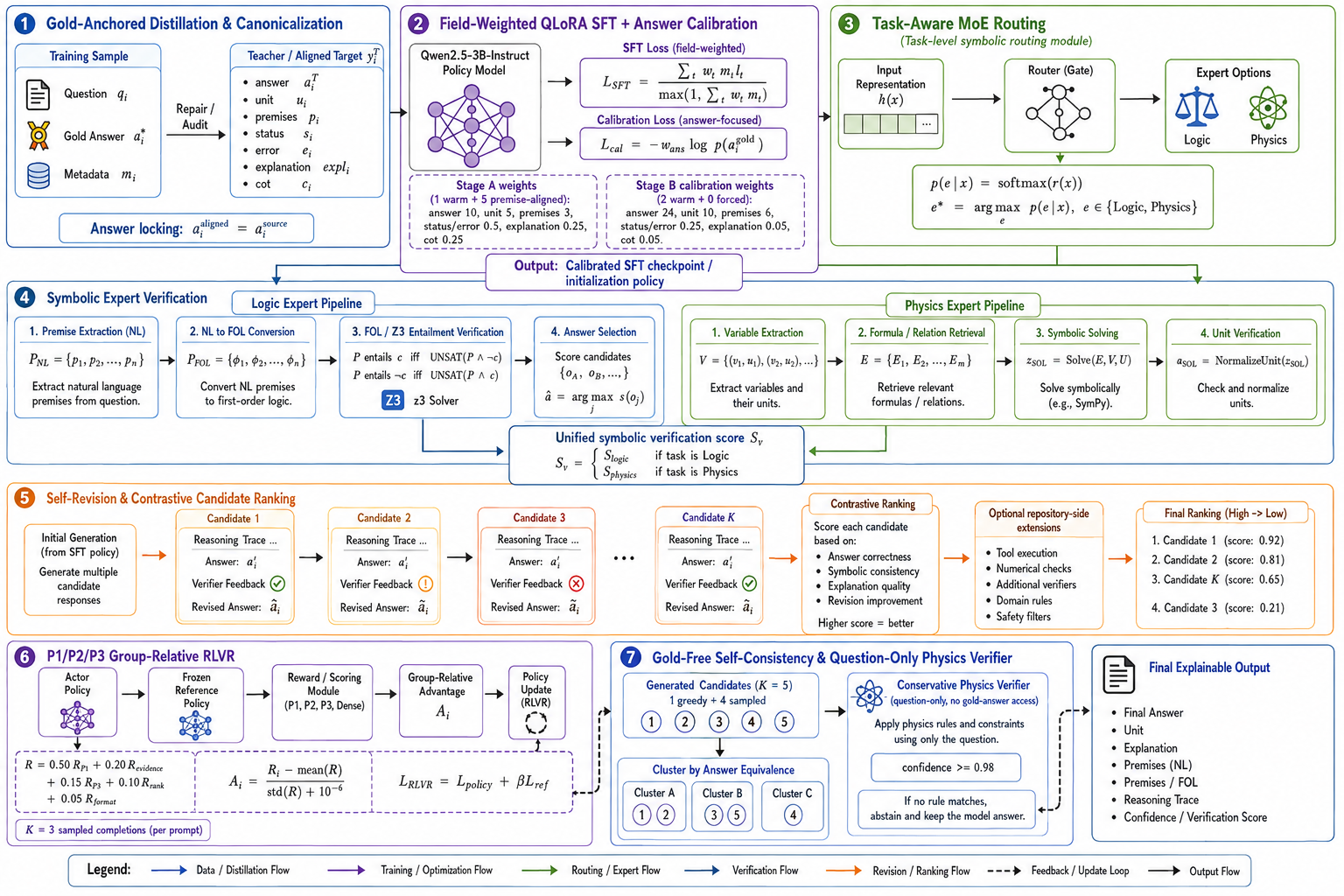}
    \caption{Overview of the proposed verifier-guided framework.
    Gold-anchored targets support QLoRA SFT and calibration. A task-aware
    MoE router selects the Logic/FOL/Z3 or Physics/SOL verification path.
    Verifier signals support revision and group-relative RLVR. At inference,
    one greedy and four sampled responses are grouped by answer equivalence
    before optional question-only physics verification.}
    \label{fig:overall_pipeline}
\end{figure*}

We formulate transparent educational QA as a
generation--verification--optimization problem in which both the final answer
and its reasoning are evaluated. As shown in Fig.~\ref{fig:overall_pipeline},
the framework consists of seven stages:
\textbf{(1)} gold-anchored canonicalization,
\textbf{(2)} field-weighted QLoRA SFT and calibration,
\textbf{(3)} task-aware Mixture-of-Experts (MoE) routing,
\textbf{(4)} symbolic verification,
\textbf{(5)} self-revision and candidate ranking,
\textbf{(6)} P1/P2/P3 group-relative RLVR, and
\textbf{(7)} gold-free self-consistency with optional physics verification.

Our MoE is a \emph{task-level routing mechanism}, not a Transformer-internal
sparse MoE architecture \cite{fedus2022switch,jiang2024mixtral}. The LLM
backbone is unchanged; the router only selects an external Logic or Physics
verification expert.

\subsection{Gold-Anchored QLoRA Fine-Tuning}

For each example
$x_i=(q_i,a_i^{*},m_i)$, where $q_i$ is the question,
$a_i^{*}$ is the authoritative answer, and $m_i$ contains task metadata,
teacher-generated supervision may be repaired while the source answer
remains locked:

\begin{equation}
\begin{aligned}
\mathmakebox[1.65cm][l]{y_i^{T}}
&= \operatorname{Repair}(x_i,a_i^{*}),\\
\mathmakebox[1.65cm][l]{a_i^{\mathrm{aligned}}}
&= a_i^{*}.
\end{aligned}
\label{eq:gold_anchor}
\end{equation}

The aligned target contains the answer, unit, premises, explanation, and a
compact reasoning trace, preventing noisy teacher outputs from replacing
authoritative labels.

We adapt Qwen2.5-3B-Instruct \cite{qwen25} using 4-bit NF4 QLoRA
\cite{dettmers2023qlora}, following LoRA \cite{hu2021lora}, with
$r=32$, $\alpha=64$, and dropout $0.05$. Completion-only SFT uses

\begin{equation}
\mathcal{L}_{\mathrm{SFT}}
=
\frac{\sum_t w_t m_t \ell_t}
     {\max\!\left(1,\sum_t w_t m_t\right)},
\label{eq:sft}
\end{equation}

where $m_t$ masks prompt tokens and $w_t$ assigns larger weights to answer,
unit, and premise fields. A subsequent answer-focused calibration stage
further strengthens these fields before RLVR.

\subsection{Task-Aware Mixture-of-Experts Routing and Symbolic Verification}

Unlike neural MoE models that route tokens through trainable expert layers,
our task-aware MoE operates only at the verification level. Given an input
representation $h(x)$, the router estimates

\begin{equation}
p(e\mid x)
=
\operatorname{softmax}
\!\left(W_r h(x)+b_r\right),
\label{eq:router_prob}
\end{equation}

and selects

\begin{equation}
e^{*}
=
\arg\max_{e} p(e\mid x),
\quad
e\in\{\mathrm{Logic},\mathrm{Physics}\}.
\label{eq:router_select}
\end{equation}

Thus, routing does not modify the Qwen Transformer or activate
Transformer-level experts; it only dispatches each problem to an external
task-specific verifier.

\noindent\textbf{\textit{1) Logic Expert.}} For logic tasks, natural-language premises are converted into
first-order-logic (FOL) representations and checked using Z3
\cite{demoura2008z3}. A candidate claim $c$ is entailed when

\begin{equation}
P\models c
\iff
\operatorname{UNSAT}(P\land\neg c).
\label{eq:logic}
\end{equation}

Contradiction is checked analogously through
$\operatorname{UNSAT}(P\land c)$. For multiple-choice questions, candidate
options are compared according to their symbolic support.

\noindent\textbf{\textit{2) Physics Expert.}} For physics tasks, the verifier extracts variables and units, retrieves a
supported relation set $\mathcal{E}$, solves the relation, and normalizes the
result:

\begin{equation}
\begin{aligned}
\mathmakebox[1.55cm][l]{z_{\mathrm{SOL}}}
&= \operatorname{Solve}(\mathcal{E},V,U),\\
\mathmakebox[1.55cm][l]{a_{\mathrm{SOL}}}
&= \operatorname{NormalizeUnit}(z_{\mathrm{SOL}}).
\end{aligned}
\label{eq:physics}
\end{equation}

The physics expert checks formula applicability, numerical consistency, and
units, while the LLM remains the primary reasoning model.

\subsection{Verifier-Guided Self-Revision}

Verifier feedback identifies unsupported premises, incorrect formulas, unit
inconsistencies, or answer--explanation conflicts. A candidate may therefore
be revised as

\begin{equation}
\tilde{y}_k
=
\operatorname{Revise}
\!\left(
y_k,
F_{\mathrm{verifier}}(y_k)
\right),
\label{eq:revision}
\end{equation}

where $F_{\mathrm{verifier}}$ denotes task-specific feedback. Candidates are
also compared using verifier-derived quality signals, encouraging responses
whose answers and reasoning are jointly better supported.

\subsection{P1/P2/P3 Group-Relative RLVR}

Starting from the calibrated SFT checkpoint, we apply reinforcement learning
with verifiable rewards (RLVR) following One-Shot-RLVR
\cite{wang2025oneshotrlvr} and the group-relative optimization principle of
GRPO \cite{shao2024deepseekmath}. For each training question, $K=3$
responses are sampled.

The reward used in the final experiment is

\begin{equation}
\small
\begin{aligned}
R_{i,k} ={}& 0.50R_{\mathrm{exact}}+0.20R_{\mathrm{dense}}\\
           &+0.15R_{\mathrm{P2}}+0.10R_{\mathrm{P3}}
            +0.05R_{\mathrm{fmt}}.
\end{aligned}
\label{eq:reward}
\end{equation}

Here, $R_{\mathrm{exact}}$ measures exact final-answer correctness,
$R_{\mathrm{dense}}$ provides task-aware partial answer credit,
$R_{\mathrm{P2}}$ measures premise consistency for logic or unit consistency
for physics, and $R_{\mathrm{P3}}$ rewards explicit reasoning depth:

\begin{equation}
R_{\mathrm{P3}}
=
\min
\left(
1,\frac{N_{\mathrm{CoT}}}{4}
\right).
\label{eq:p3_reward}
\end{equation}

For each response group, the relative advantage is

\begin{equation}
A_{i,k}
=
\frac{R_{i,k}-\mu_i}
     {\sigma_i+\epsilon},
\label{eq:advantage}
\end{equation}

with

\begin{equation}
\mu_i
=
\frac{1}{K}
\sum_{k=1}^{K}R_{i,k}.
\label{eq:reward_mean}
\end{equation}

Groups with negligible reward variance are skipped because they provide no
useful relative preference signal.

A frozen copy of the calibrated SFT checkpoint serves as the reference
policy. In the active implementation, reference regularization uses the
difference between sequence-average log probabilities:

\begin{equation}
\mathcal{L}_{\mathrm{ref}}
=
\left(
\bar{\ell}_{\theta}
-
\bar{\ell}_{\mathrm{ref}}
\right)^2.
\label{eq:ref}
\end{equation}

Following the group-relative RL formulation \cite{shao2024deepseekmath} and
verifiable-reward training setting \cite{wang2025oneshotrlvr}, the policy
uses the clipped surrogate implemented in our training code. Let
$\bar{\ell}_{\theta}$ and $\bar{\ell}_{\mathrm{old}}$ denote the
sequence-average log probabilities under the current and rollout policies:

\begin{equation}
\rho_{i,k}(\theta)
=
\exp\!\left(
\bar{\ell}_{\theta,i,k}
-
\bar{\ell}_{\mathrm{old},i,k}
\right).
\label{eq:ratio}
\end{equation}

The group-relative policy loss is

\begin{equation}
\begin{aligned}
\mathcal{L}_{\mathrm{policy}}
=-\frac{1}{K}\sum_{k=1}^{K}\min\Big(&
\rho_{i,k}A_{i,k},\\[-1mm]
&\operatorname{clip}(\rho_{i,k},1-\varepsilon,1+\varepsilon)A_{i,k}
\Big),
\end{aligned}
\label{eq:policy_loss}
\end{equation}

where $\varepsilon=0.2$ in the final run. Combining this clipped objective
with reference regularization gives the RLVR loss

\begin{equation}
\mathcal{L}_{\mathrm{RLVR}}
=
\mathcal{L}_{\mathrm{policy}}
+
\beta\mathcal{L}_{\mathrm{ref}}.
\label{eq:rlvr}
\end{equation}

Thus, $A_{i,k}$ favors candidates whose verifiable reward exceeds the group
mean, clipping limits overly large policy updates, and
$\mathcal{L}_{\mathrm{ref}}$ constrains drift from the calibrated SFT policy.

\subsection{Gold-Free Self-Consistency and Final Output}

At held-out inference, five responses are generated,

\begin{equation}
\mathcal{Y}
=
\left\{
y^{(g)},
y^{(1)},
y^{(2)},
y^{(3)},
y^{(4)}
\right\},
\label{eq:candidates}
\end{equation}

where $y^{(g)}$ is generated greedily and the remaining four responses are
sampled. They are grouped by task-aware answer equivalence:

\begin{equation}
\mathcal{C}
=
\operatorname{ClusterEquivalent}(\mathcal{Y}).
\label{eq:cluster}
\end{equation}

The largest cluster is selected. If clusters have equal size, preference is
given to the cluster containing the greedy response; remaining ties are
resolved by a gold-free structural quality score
\cite{wang2023selfconsistency}. No validation gold label is used during this
selection.

For physics, a conservative question-only verifier may override the selected
answer only when

\begin{equation}
c_{\mathrm{verifier}}\geq0.98,
\label{eq:confidence}
\end{equation}

otherwise it abstains. The final output contains the answer and, where
applicable, unit, supporting premises, symbolic/FOL information, reasoning
trace, and verification confidence.

\section{Experiments}
\label{sec:experiments}

\subsection{Experimental Setup}

We evaluate the proposed framework on the EXACT logic and physics data
\cite{exact2026}. After preprocessing, 2,162 examples are divided using a
group-safe 80:20 split into 1,724 training and 438 held-out validation
samples. Shared logic premise blocks, duplicate normalized physics questions,
and identical model inputs are kept within the same partition to reduce
leakage. Retrieval is disabled (\texttt{RETRIEVAL\_TOP\_K=0}), and validation
labels are accessed only after generation.

Table~\ref{tab:dataset} summarizes the resulting data split.

\begin{table}[H]
\centering
\caption{Experimental data composition.}
\label{tab:dataset}
\scriptsize
\setlength{\tabcolsep}{4pt}
\begin{tabular}{lrrr}
\hline
\textbf{Task} & \textbf{Total} & \textbf{Train} & \textbf{Val.} \\
\hline
Logic Multiple Choice & 360   & 287   & 73  \\
Logic Yes/No          & 415   & 331   & 84  \\
Logic Uncertain       & 33    & 26    & 7   \\
Physics               & 1,354 & 1,080 & 274 \\
\hline
\textbf{Total}        & \textbf{2,162} &
\textbf{1,724} & \textbf{438} \\
\hline
\end{tabular}
\end{table}

The backbone is Qwen2.5-3B-Instruct \cite{qwen25} with 4-bit NF4 QLoRA
\cite{dettmers2023qlora}. Maximum sequence, prompt, and generation lengths
are 1,280, 1,024, and 224 tokens. Stage A performs three SFT epochs at
$5\times10^{-5}$, Stage B performs one calibration epoch at
$2\times10^{-5}$, and Stage C performs 240 RLVR rollout steps at
$2\times10^{-6}$.

Evaluation uses \textbf{five-generation gold-free self-consistency}: one
greedy response and four sampled responses ($T=0.35$, top-$p=0.90$). The
question-only physics verifier intervenes only for supported rules with
confidence $\geq0.98$. Experiments run on one NVIDIA L4 24\,GB GPU in a
PyTorch/Hugging Face environment using Transformers, PEFT, Accelerate,
bitsandbytes, and repository-side FOL/Z3 and SOL modules.

\subsection{Quantitative Results}

We report three complementary reasoning dimensions. \textbf{P1} is final
answer correctness after the complete inference pipeline. \textbf{P2} is a
task-aware supporting-consistency score: for logic it is the partial F1
between predicted and reference premise indices, whereas for physics it checks
normalized unit consistency (with partial multi-unit credit for multi-valued
answers). \textbf{P3} is the generated reasoning-depth proxy defined in
Eq.~\eqref{eq:p3_reward}. Table~\ref{tab:overall_results} compares calibrated
SFT and the final RLVR policy on the same 438 held-out examples.

\begin{table*}[!t]
\centering
\caption{Held-out validation results (\%). Macro P1 is the unweighted mean
over the four evaluated task families.}
\label{tab:overall_results}
\scriptsize
\setlength{\tabcolsep}{5pt}
\begin{tabular}{lrrrrrrrr}
\hline
\textbf{Model} &
\textbf{P1} &
\textbf{Macro P1} &
\textbf{Logic P1} &
\textbf{Physics P1} &
\textbf{P2} &
\textbf{P3} &
\textbf{Format} &
\textbf{Strict} \\
\hline
Calibrated SFT
& \textbf{56.62}
& \textbf{76.82}
& \textbf{83.54}
& \textbf{40.51}
& \textbf{77.70}
& 50.68
& \textbf{98.17}
& \textbf{41.44} \\

RLVR
& 55.94
& 76.14
& 82.32
& 40.15
& 75.33
& \textbf{72.20}
& 94.52
& 40.14 \\
\hline
$\Delta$
& -0.68
& -0.69
& -1.22
& -0.36
& -2.37
& \textbf{+21.52}
& -3.65
& -1.30 \\
\hline
\end{tabular}
\end{table*}

The three metrics reveal different effects of RLVR. \textbf{P1} changes only
from 56.62\% to 55.94\% ($-0.68$ points), so the completed RLVR run does not
produce an additional held-out answer-accuracy gain. \textbf{P2} decreases
from 77.70\% to 75.33\% ($-2.37$ points), indicating that richer generated
reasoning does not automatically preserve premise selection or physical-unit
consistency. In contrast, \textbf{P3} rises from 50.68\% to 72.20\%, a gain
of 21.52 points, making reasoning depth the dominant observed post-training
change.

This pattern is technically consistent with the group-relative reward. Exact
P1 has the largest individual coefficient, but candidates inside a rollout
group frequently share the same answer reward. When this occurs, P1 contributes
little relative ranking signal, whereas differences in P2, P3, dense answer
credit, and formatting can still produce non-zero advantages. Since P3 is
explicitly rewarded, the policy can therefore learn to emit more complete
reasoning traces even when final-answer accuracy has already saturated within
a group. The format decrease from 98.17\% to 94.52\% further shows that
longer structured outputs create more opportunities for malformed fields or
schema violations.

P2 should also be interpreted jointly with the task mixture rather than as a
single homogeneous quantity. Physics contributes 274 of the 438 validation
examples and retains high unit consistency (91.79\% for SFT and 89.23\% for
RLVR), which lifts the micro-averaged P2. Logic P2 instead measures premise
selection and is substantially lower. Consequently, the overall P2 decrease
reflects both a modest physics-unit decline and weaker premise consistency in
some logic categories, rather than one uniform failure mode.

Finally, P3 is a reasoning-depth proxy, not a proof of logical validity. A
longer trace can make intermediate assumptions easier to inspect while still
containing an unsupported premise or numerical mistake. Reporting P1, P2,
P3, and Strict together therefore separates final correctness, supporting
consistency, reasoning explicitness, and end-to-end structural validity. The
small Macro-P1 change (76.82\% to 76.14\%) likewise indicates that the P1
shift is not caused by collapse of a single task family.

\noindent\textbf{\textit{1) Task-Level P1/P2/P3 Analysis.}}

Table~\ref{tab:task_results} decomposes all three metrics by task family. The
P2 values are computed with the same task-aware definition used in the overall
score and are therefore directly consistent with Table~\ref{tab:overall_results}.

\begin{table*}[!t]
\centering
\begin{minipage}[t]{0.36\textwidth}
\centering
\captionof{table}{Task-level P1/P2/P3 performance (\%).}
\label{tab:task_results}
\tiny
\setlength{\tabcolsep}{1.0pt}
\begin{tabular}{lrrrrrr}
\hline
& \multicolumn{2}{c}{\textbf{P1}} & \multicolumn{2}{c}{\textbf{P2}} & \multicolumn{2}{c}{\textbf{P3}} \\
\textbf{Task} & \textbf{SFT} & \textbf{RLVR} & \textbf{SFT} & \textbf{RLVR} & \textbf{SFT} & \textbf{RLVR} \\
\hline
Logic MC & 91.78 & 91.78 & 57.26 & 53.70 & 40.75 & \textbf{72.95} \\
Logic Yes/No & 75.00 & 72.62 & 52.20 & 51.87 & 42.86 & \textbf{75.00} \\
Logic Uncertain & 100.00 & 100.00 & 45.31 & 38.16 & 53.57 & \textbf{71.43} \\
Physics & 40.51 & 40.15 & 91.79 & 89.23 & 55.66 & \textbf{71.17} \\
\hline
\end{tabular}
\end{minipage}
\hfill
\begin{minipage}[t]{0.39\textwidth}
\centering
\captionof{table}{Inference-component ablation on P1 (\%).}
\label{tab:ablation}
\tiny
\setlength{\tabcolsep}{1.35pt}
\begin{tabular}{l p{2.35cm} rrr}
\hline
\textbf{Policy} & \textbf{Inference} & \textbf{Overall} & \textbf{Logic} & \textbf{Physics} \\
\hline
SFT & Greedy & 50.46 & 83.54 & 30.66 \\
SFT & + Five-generation self-consistency & 51.14 & 83.54 & 31.75 \\
SFT & + Physics verifier & \textbf{56.62} & 83.54 & \textbf{40.51} \\
\hline
RLVR & Greedy & 48.86 & 82.32 & 28.83 \\
RLVR & + Five-generation self-consistency & 50.23 & 82.32 & 31.02 \\
RLVR & + Physics verifier & \textbf{55.94} & 82.32 & \textbf{40.15} \\
\hline
\end{tabular}
\end{minipage}
\hfill
\begin{minipage}[t]{0.23\textwidth}
\centering
\captionof{table}{Physics P1 by answer representation (\%).}
\label{tab:physics_types}
\tiny
\setlength{\tabcolsep}{1.8pt}
\begin{tabular}{lrrr}
\hline
\textbf{Type} & \textbf{N} & \textbf{SFT} & \textbf{RLVR} \\
\hline
Plain numeric & 195 & 49.74 & 49.74 \\
Scientific numeric & 48 & 6.25 & 6.25 \\
Binary & 4 & 75.00 & 75.00 \\
Categorical text & 8 & 37.50 & 37.50 \\
Numeric assignment & 3 & 66.67 & 66.67 \\
Multiple values & 6 & 50.00 & 33.33 \\
\hline
\end{tabular}
\end{minipage}
\end{table*}

\begingroup\scriptsize
For \textbf{P1}, Logic Multiple Choice remains exactly 91.78\%, Logic
Uncertain remains 100\% on its seven-example subset, Logic Yes/No decreases
by 2.38 points, and Physics changes only from 40.51\% to 40.15\%. Hence, the
small global P1 decrease is distributed across limited task-level changes
rather than a broad loss of answer capability.

For \textbf{P2}, Logic Multiple Choice changes from 57.26\% to 53.70\%,
Logic Yes/No from 52.20\% to 51.87\%, Logic Uncertain from 45.31\% to
38.16\%, and Physics from 91.79\% to 89.23\%. The largest percentage-point
logic decrease occurs on the seven-example Uncertain subset and should not be
over-interpreted. The physics P2 value is much higher because it measures unit
consistency, whereas logic P2 is a partial premise-selection F1. This confirms
that P2 is best used as a task-aware diagnostic rather than as a substitute for
P1.

For \textbf{P3}, every populated task family improves: Logic Multiple Choice
increases by 32.19 points, Logic Yes/No by 32.14, Logic Uncertain by 17.86,
and Physics by 15.51. This consistency across heterogeneous tasks strengthens
the interpretation that RLVR primarily reshapes reasoning explicitness. The
physics result is especially informative: P3 rises substantially while P1
stays near 40\%, showing that producing a deeper trace is not sufficient to
solve numerical execution or formula-selection errors.

\subsection{Ablation Study}

We isolate the contribution of greedy decoding, five-generation gold-free
self-consistency, and deterministic physics verification. Table~\ref{tab:ablation}
reports P1 because this ablation is designed specifically to measure final
answer correction at inference; complete component-wise P2/P3 traces are not
logged for every intermediate decoding mode and are therefore not inferred.

Five-generation self-consistency increases overall P1 by 0.68 points for SFT
and 1.37 points for RLVR. Symbolic verification provides a substantially
larger physics gain: relative to self-consistency, physics P1 improves by
8.76 points for SFT and 9.12 points for RLVR. The verifier intervenes on only
43 of 274 physics queries (15.69\%), correcting 26 SFT and 27 RLVR errors
while introducing two regressions for each policy. This sparse intervention
pattern supports the intended design: the symbolic module acts as a selective
checker for supported high-confidence relations rather than as a second model
that replaces the LLM on every physics query.

The ablation also clarifies where the observed system-level gain originates.
For both policies, moving from greedy decoding to five-generation
self-consistency yields only a small increase, whereas the physics verifier
produces the dominant final correction. This separation is important because
it prevents the verifier gain from being misattributed to RLVR itself: the
policy changes the distribution and structure of generated reasoning, while
the deterministic checker repairs a smaller subset of supported numerical
cases after generation. The similar verifier gains for SFT and RLVR further
suggest that the symbolic module is exploiting task structure that remains
useful across both neural checkpoints rather than compensating for one
particular policy.

\noindent\textbf{\textit{1) Physics Error Analysis.}}

Table~\ref{tab:physics_types} further decomposes physics answer correctness by
representation. Plain numeric answers reach 49.74\% P1, whereas
scientific-notation answers reach only 6.25\%. At the same time, physics P2
(unit consistency) remains 89.23\% after RLVR. The gap between high unit
consistency and much lower answer correctness indicates that the dominant
physics failures are not primarily unit recognition; they are more consistent
with formula selection, numerical execution, exponent normalization, and
answer representation errors.

The RLVR training trajectory provides a complementary explanation. Of 240 rollout
groups, 57 (23.75\%) have zero reward variance and are skipped, leaving 183
informative groups and 30 optimizer updates. The right panel of
Fig.~\ref{fig:training_dynamics} summarizes 20-step rolling means observed
during RLVR training. The first versus last 20-step means are approximately
0.86$\rightarrow$0.90 for reward, 0.93$\rightarrow$0.95 for P1,
0.67$\rightarrow$0.75 for P2, and 0.53$\rightarrow$0.74 for P3. Thus, the
value near 0.94 refers to sampled P1 during RLVR training rather than the
held-out P1 values in Tables~\ref{tab:overall_results}--\ref{tab:ablation}.
Under Eq.~\eqref{eq:advantage}, zero-variance groups provide no relative
ranking signal, so optimization is concentrated on groups where candidate
rewards differ; the strongest sustained change is consequently observed in P3.

\subsection{Comparison with Prior Reasoning Frameworks}

Table~\ref{tab:prior_benchmark} provides a contextual comparison with
representative reasoning and verification approaches. Because datasets,
models, inference budgets, and evaluation metrics differ, the reported
numbers are not directly comparable.

\begin{table*}[!t]
\centering
\caption{Contextual comparison with representative reasoning and verification approaches. Results are not directly comparable across benchmarks.}
\label{tab:prior_benchmark}
\scriptsize
\setlength{\tabcolsep}{3.3pt}
\begin{tabular}{p{2.6cm}p{2.8cm}p{3.0cm}p{4.0cm}p{2.8cm}}
\hline
\textbf{Method} & \textbf{Benchmark / Model} & \textbf{Supervision} & \textbf{Reasoning / Verification} & \textbf{Reported Result} \\
\hline
Let's Verify Step by Step \cite{lightman2023verify} & MATH subset / large LLM & Process supervision & Process Reward Model & 78.2\% accuracy \\
DeepSeekMath \cite{shao2024deepseekmath} & MATH / DeepSeekMath-7B & SFT + GRPO & Outcome-verifiable group-relative optimization & 51.7\%; 60.9\% with 64-sample self-consistency \\
One-Shot-RLVR \cite{wang2025oneshotrlvr} & MATH500 / Qwen2.5-Math-1.5B & One-example RLVR & Group-relative verifiable reward & 36.0\% $\rightarrow$ 73.6\% \\
Scientific Logicality \cite{yu2026scientificlogicality} & PhysLogic / multiple LLMs & Logicality-guided & Process-aware scientific reasoning evaluation & Improved logicality and physics performance \\
\hline
\textbf{Ours} & EXACT / Qwen2.5-3B-Instruct & Gold-anchored QLoRA + RLVR & FOL/Z3 + Physics/SOL + five-generation gold-free self-consistency & \textbf{55.94\% P1}; \textbf{75.33\% P2}; \textbf{72.20\% P3} \\
\hline
\end{tabular}
\end{table*}

Prior work emphasizes either intermediate process supervision or verifiable
mathematical outcomes. Our framework instead combines heterogeneous logic and
physics reasoning with task-specific symbolic verification and reports all
three core diagnostics together. P1 captures whether the final answer is
correct, P2 measures whether its supporting premises or physical units remain
consistent, and P3 measures whether the generated response exposes explicit
intermediate reasoning. The comparison is therefore methodological rather
than a direct leaderboard.

\subsection{Discussion}

The verified results expose a clear three-way P1--P2--P3 trade-off. Relative
to calibrated SFT, RLVR changes P1 by $-0.68$ points and P2 by $-2.37$ points
but improves P3 by $+21.52$ points. The result does not support the claim that
RLVR improves every metric; instead, it supports a more specific conclusion:
under the present reward weighting and optimization budget, RLVR mainly
strengthens reasoning explicitness while preserving answer accuracy within
approximately one percentage point of the calibrated baseline.

P2 provides an additional diagnostic that prevents the P3 result from being
overstated. The modest P2 decline shows that a deeper response can introduce
an incorrect or unnecessary premise, or lose unit consistency, even when the
trace becomes more complete. Conversely, P1 can remain correct despite a
weaker supporting trace. Reporting all three metrics therefore makes the
source of a gain or degradation visible and better matches the objective of
transparent educational QA than relying on one aggregate accuracy number.

Five-generation self-consistency and symbolic verification address different
parts of this trade-off. Self-consistency reduces single-generation variance,
whereas the deterministic physics verifier selectively corrects cases that can
be solved by supported high-confidence relations. The verifier's large physics
P1 gain despite sparse activation indicates that some residual errors are
execution failures rather than failures to produce reasoning text. Together,
these observations motivate treating the neural policy, reasoning objective,
and symbolic checker as complementary components rather than attributing all
system-level improvement to RLVR alone.

\scriptsize
\subsection{Visualization Results}

The five diagnostics summarize the same measurements reported in the tables.
Figure~\ref{fig:tasklevel_p123} shows task-level P1, P2, and P3 from
Table~\ref{tab:task_results}; Fig.~\ref{fig:training_dynamics} separates
inference-stage P1 correction from RLVR training-time dynamics.

\vspace{1pt}
\begin{center}
\includegraphics[width=0.315\columnwidth]{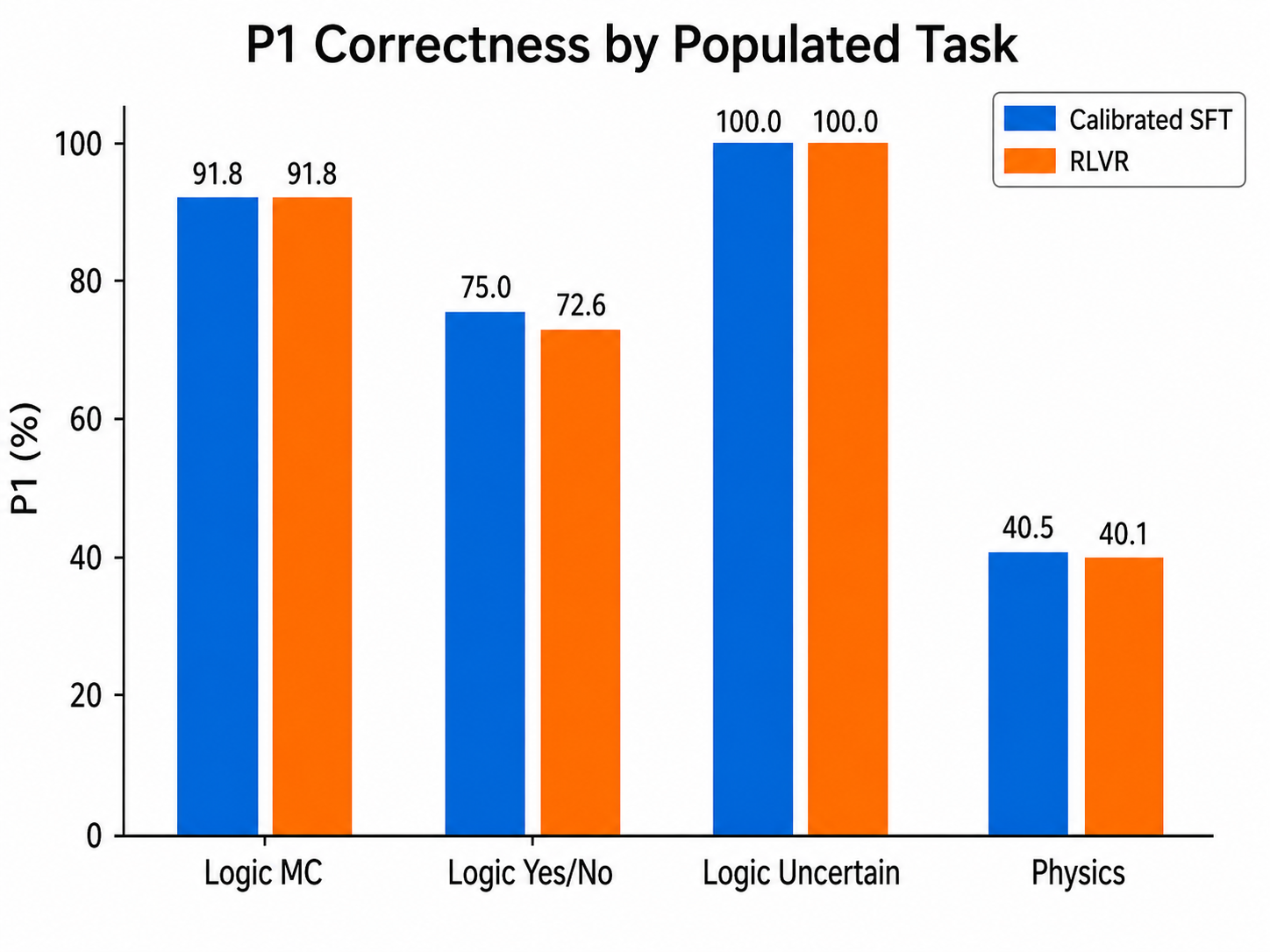}\hfill
\includegraphics[width=0.315\columnwidth]{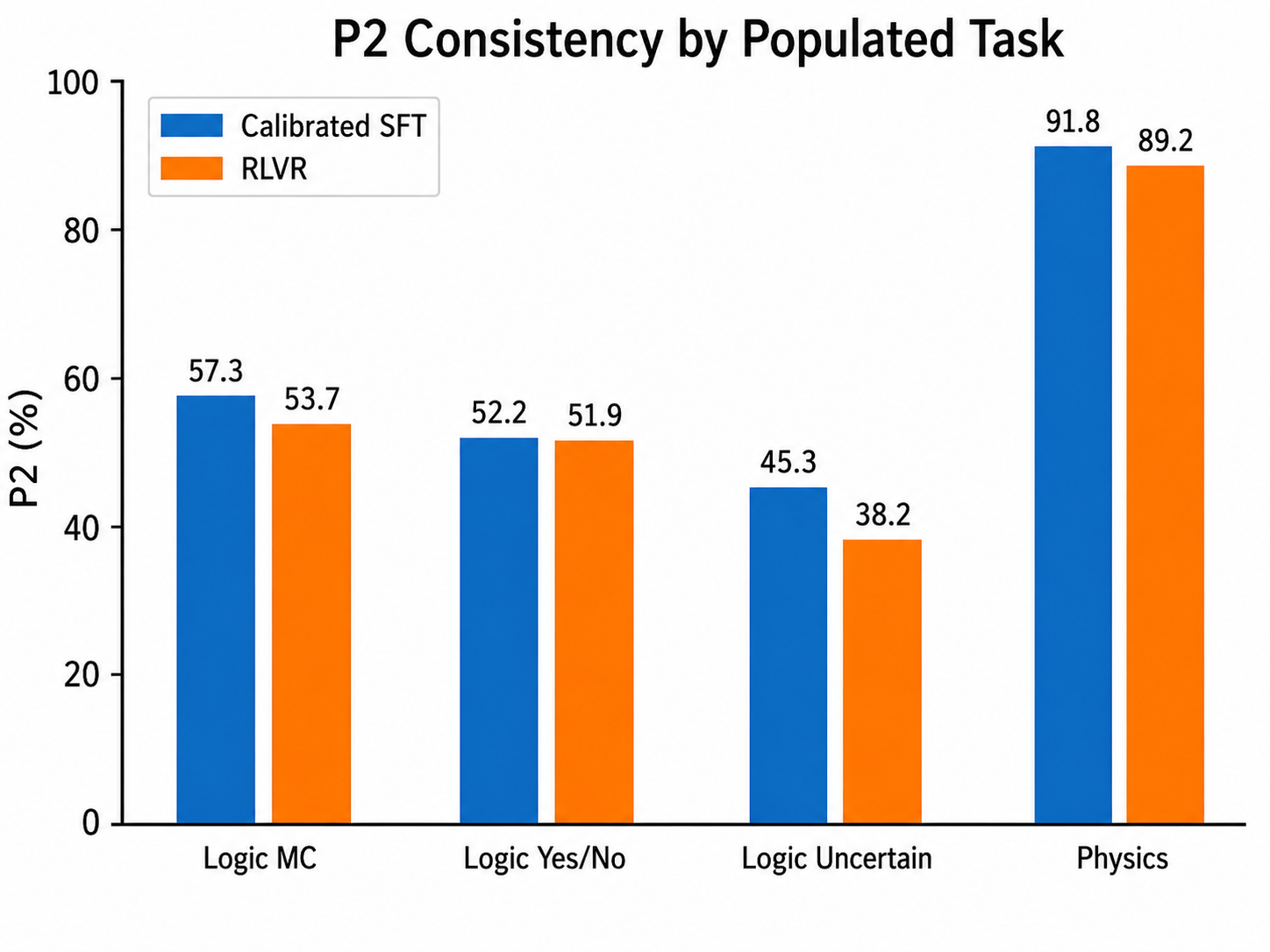}\hfill
\includegraphics[width=0.315\columnwidth]{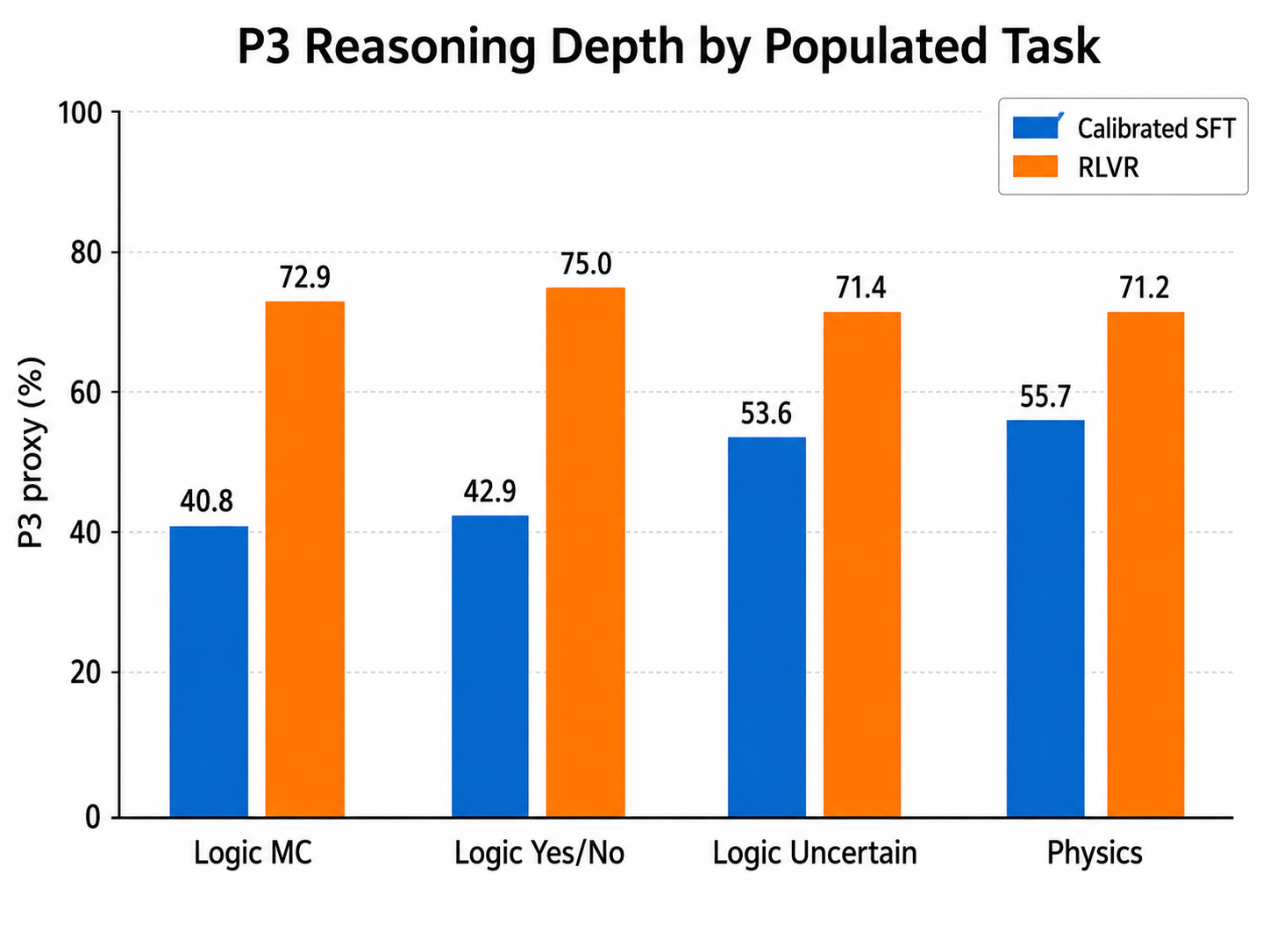}
\captionsetup{font=scriptsize,skip=1pt}
\captionof{figure}{Task-level diagnostics from Table~\ref{tab:task_results}. Left: P1 correctness. Middle: P2 consistency. Right: P3 reasoning depth.}
\label{fig:tasklevel_p123}
\end{center}
\vspace{-2pt}

Figure~\ref{fig:tasklevel_p123} makes the P1--P2--P3 trade-off directly
checkable against Table~\ref{tab:task_results}. P1 is nearly unchanged:
Logic MC remains 91.8\%, Logic Yes/No changes 75.0$\rightarrow$72.6\%, Logic
Uncertain remains 100.0\%, and Physics changes 40.5$\rightarrow$40.1\%.
P2 decreases from 57.3$\rightarrow$53.7\%, 52.2$\rightarrow$51.9\%,
45.3$\rightarrow$38.2\%, and 91.8$\rightarrow$89.2\%, respectively. In
contrast, P3 rises in every task: 40.8$\rightarrow$72.9\%,
42.9$\rightarrow$75.0\%, 53.6$\rightarrow$71.4\%, and
55.7$\rightarrow$71.2\%. The visual pattern therefore supports the same
conclusion as Table~\ref{tab:overall_results}: RLVR changes reasoning depth
more strongly than final-answer correctness.

\vspace{2pt}
\begin{center}
\includegraphics[width=0.485\columnwidth]{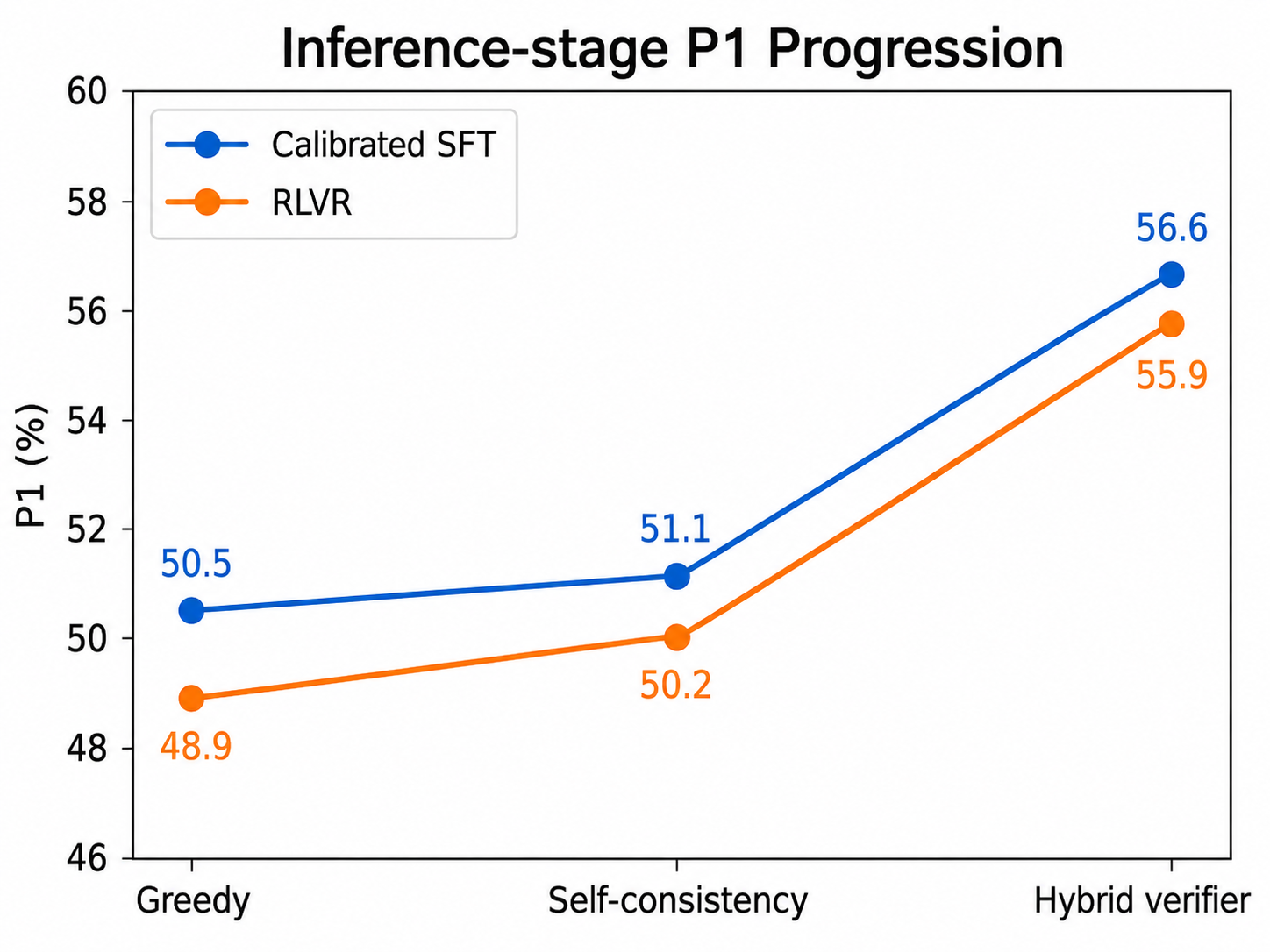}\hfill
\includegraphics[width=0.485\columnwidth]{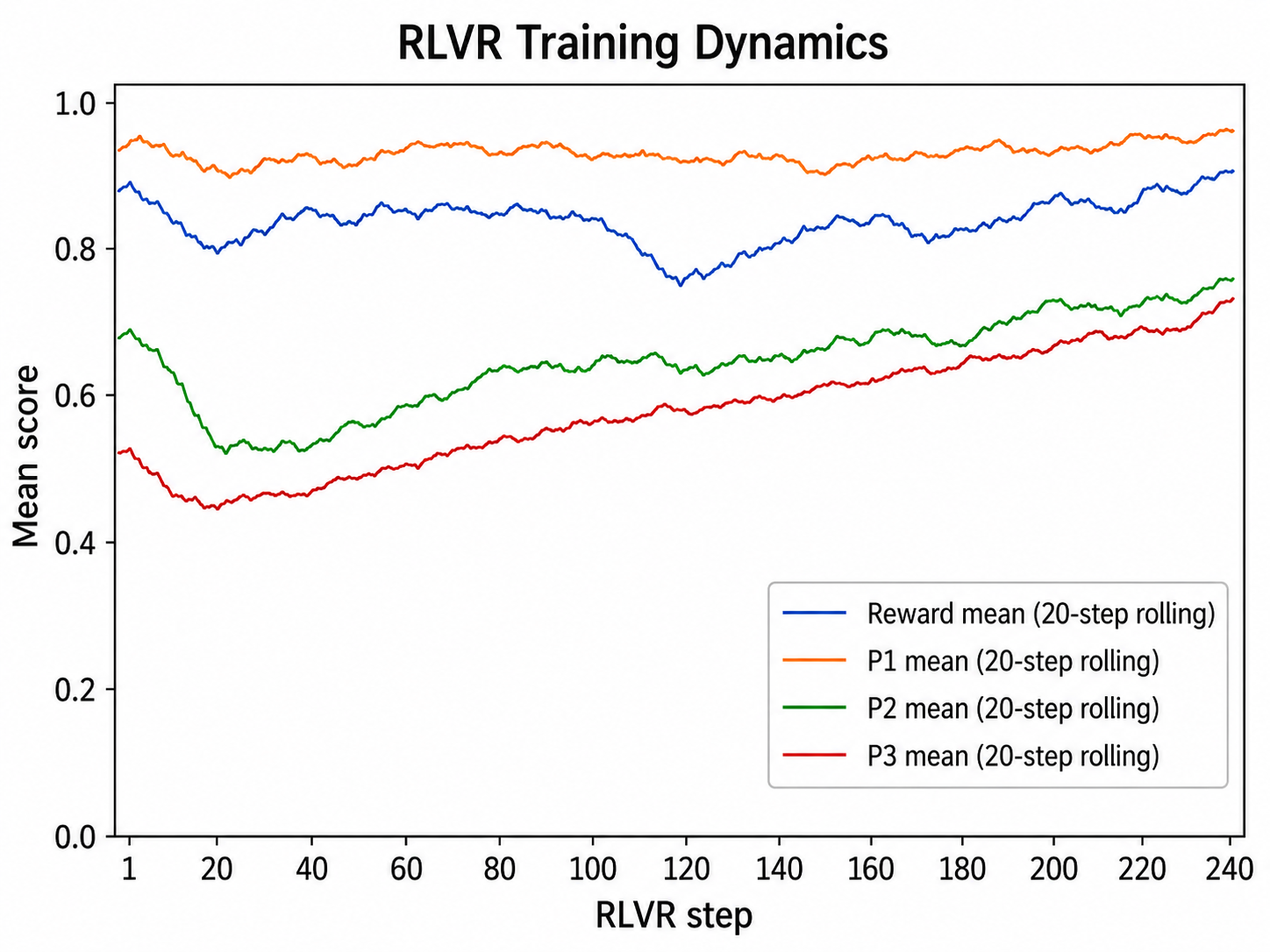}
\captionof{figure}{Inference and training dynamics. Left: P1 progression from Table~\ref{tab:ablation}. Right: 20-step rolling reward/P1/P2/P3 during RLVR training.}
\label{fig:training_dynamics}
\end{center}
\vspace{-2pt}

Figure~\ref{fig:training_dynamics} (left) reproduces Table~\ref{tab:ablation}:
SFT progresses 50.5$\rightarrow$51.1$\rightarrow$56.6\%, while RLVR progresses
48.9$\rightarrow$50.2$\rightarrow$55.9\% from greedy decoding to
self-consistency and then hybrid verification. Hence, self-consistency gives
the modest model-only gain and the verifier supplies the larger final
correction. The right panel is not a held-out accuracy plot; it summarizes
20-step rolling statistics observed during RLVR training. Its first/last
20-step means are reward 0.86/0.90, P1 0.93/0.95, P2 0.67/0.75, and P3
0.53/0.74. The value near 0.94 therefore refers to sampled training-time P1,
not the held-out values reported in Tables~\ref{tab:overall_results}
or~\ref{tab:ablation}. Together, the five plots separate RLVR's reasoning-depth
shift, self-consistency's variance reduction, and the physics verifier's
selective correctness gain.

\Needspace{0.28\textheight}
\subsection{Inference Results}

The final inference pipeline combines five-generation gold-free
self-consistency with optional symbolic verification. One greedy and four
sampled responses are grouped by answer equivalence, after which the selected
physics response may be conservatively checked by the question-only verifier.
Fig.~\ref{fig:inference_results} presents six representative correct cases
covering three logic and three physics output formats. Each case exposes the
question, premises or physical information, model answer, reasoning response,
gold answer, and verification notes. The cases are selected only after
generation for qualitative visualization; gold labels are not available to
clustering or the question-only verifier.

\begin{figure}[H]
\centering
\includegraphics[width=0.90\columnwidth]{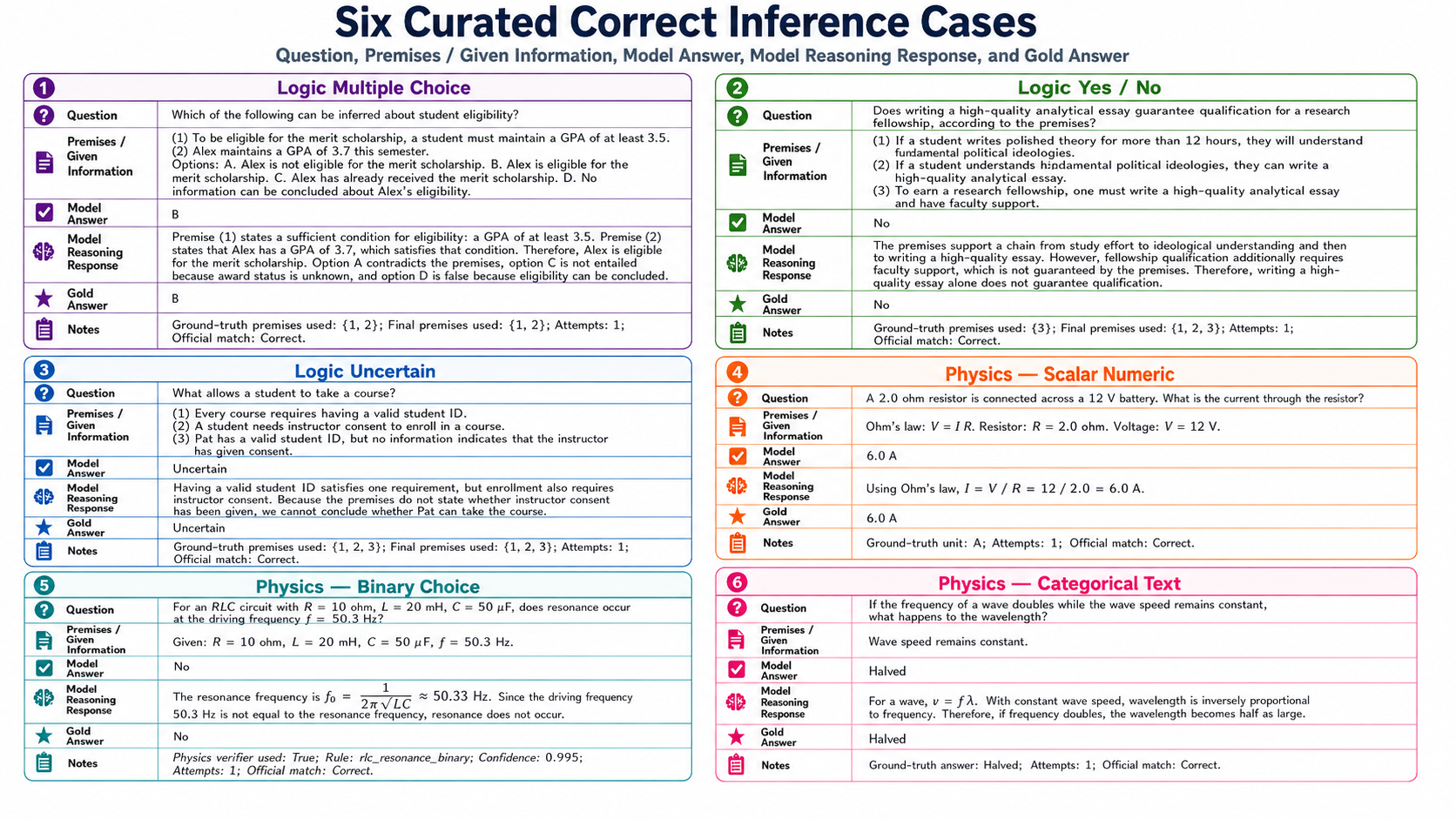}
\caption{Representative correct inference cases across six logic and physics task formats.}
\label{fig:inference_results}
\end{figure}

The qualitative cases complement the quantitative P1/P2/P3 analysis. Logic
examples make premise use directly inspectable, whereas physics examples
expose the chosen relation, numerical substitution, units, and verifier
status. They also illustrate why the metrics should not be collapsed: P1 can
be correct while premise selection is incomplete, P2 can remain high despite
a numerical answer error, and P3 can increase simply because more reasoning
steps are made explicit. The intended system therefore uses the LLM for
general reasoning and reserves symbolic intervention for supported,
verifiable cases.
\endgroup

\balance
\section{Conclusion and Future Work}
\label{sec:conclusion}
\begingroup\scriptsize

We presented a verifier-guided framework combining gold-anchored QLoRA,
task-aware symbolic routing, self-revision, and group-relative RLVR for
transparent educational QA. It separately evaluates \textbf{P1} final-answer
correctness, \textbf{P2} evidence/unit consistency, and \textbf{P3} reasoning
depth.

On 438 held-out examples, RLVR raises P3 from 50.68\% to 72.20\%, while
hybrid P1 remains close to calibrated SFT (55.94\% vs. 56.62\%) and P2 is
75.33\%. Five-generation gold-free self-consistency gives a modest model-only
gain, whereas the conservative physics verifier yields the largest physics-P1
gain. Thus, RLVR mainly reshapes explicit reasoning, while symbolic
verification selectively corrects high-confidence numerical failures.

The current RLVR objective improves reasoning depth more strongly than answer
accuracy, and longer outputs can introduce premise, unit, or formatting
inconsistencies. Physics also remains limited by formula selection, numerical
execution, and scientific-notation normalization. Future work will study
\textbf{(i)} improved P1/P2/P3 reward balancing, \textbf{(ii)} stronger
numerical and symbolic normalization, \textbf{(iii)} broader logic/physics
verifier coverage, \textbf{(iv)} complete reward-component and routing
ablations, and \textbf{(v)} larger scientific reasoning benchmarks. We also
plan to calibrate verifier confidence, use symbolic feedback earlier in
candidate revision, and test robustness under paraphrased premises, altered
numerical scales, and unit conversions, aiming to preserve the P3 gain while
improving correctness, consistency, and transfer.

Beyond these extensions, a stronger evaluation protocol should quantify the stability of the observed gains across random seeds and alternative validation splits. In particular, the seven-example Logic Uncertain subset is too small to support strong conclusions by itself, while physics dominates the held-out distribution and therefore has a disproportionate effect on overall P1. Future experiments will report confidence intervals, per-task calibration, and repeated runs so that changes in P1, P2, and P3 can be distinguished from sampling variation. We will also evaluate the verifier as an explicit selective-prediction component by measuring coverage, correction rate, regression rate, and confidence calibration rather than accuracy alone.

A second direction concerns efficiency and transfer. The current pipeline uses multiple generations for gold-free self-consistency and invokes symbolic verification only on supported high-confidence cases. We plan to study whether candidate count can be reduced adaptively when answer clusters agree early, and whether verifier routing can be learned with lower inference overhead. Finally, the same separation between answer correctness, supporting consistency, and reasoning depth can be tested on broader STEM domains and multilingual educational QA. Such experiments would clarify whether the present neural--symbolic design transfers beyond the EXACT setting while preserving the interpretability advantages of structured outputs and selective verification.

Finally, we will study reward-weight sensitivity, reference regularization, verifier-confidence calibration, repeated-run uncertainty, and the latency/memory overhead of multi-candidate generation and symbolic checking so that reasoning gains can be assessed jointly with correctness and deployment cost.
\endgroup

\begingroup
\scriptsize
\bibliographystyle{IEEEtran}
\bibliography{REFERENCES}
\endgroup

\end{document}